\documentclass[letterpaper, 10 pt, conference]{ieeeconf} 

\IEEEoverridecommandlockouts 

\usepackage{amsmath} 
\usepackage{amssymb} 
\usepackage{cite}
\usepackage{flushend}
\usepackage{tikz}
\usepackage{algorithm}
\usepackage{algpseudocode}
\usepackage[font=small,skip=1pt]{caption}
\usetikzlibrary{shapes,arrows} 
\usetikzlibrary{calc}
\tikzset{
-|-/.style={
to path={
(\tikztostart) -| ($(\tikztostart)!#1!(\tikztotarget)$) |- (\tikztotarget)
\tikztonodes
}
},
-|-/.default=0.5,
|-|/.style={
to path={
(\tikztostart) |- ($(\tikztostart)!#1!(\tikztotarget)$) -| (\tikztotarget)
\tikztonodes
}
},
|-|/.default=0.5,
}

\title{\LARGE \bf
Multi-Volume High Resolution RGB-D Mapping with Dynamic Volume Placement 
}

\author{Michael Salvato$^{1}$, Ross Finman $^{1}$, and John J. Leonard$^{1}$
\thanks{This work was partially supported by the Office of Naval Research
under grants N00014-10-1-0936, N00014-11-1-0688 and N00014-13-1-0588
and by the National Science Foundation under grant IIS-1318392, which
we gratefully acknowledge.}%
\thanks{$^{1}$M. Salvato, R. Finman, and J. J. Leonard are with the Computer Science and Artificial Intelligence Laboratory (CSAIL), Massachusetts Institute of Technology (MIT), Cambridge, MA 02139, USA. {\tt\small \{msalvato, rfinman, jleonard\}@mit.edu}}
}

\begin{document}

\maketitle
\thispagestyle{empty}
\pagestyle{empty}

\section{ABSTRACT}

\begin{abstract}
We present a novel RGB-D mapping system for generating 3D maps over spatially extended regions with higher resolution than current methods using multiple, dynamically placed mapping volumes. Our method takes in RGB-D frames and dynamically assigns multiple mapping volumes to the environment, exchanging mapping volumes between the CPU and GPU. Mapping volumes are added or removed as needed to allow for spatially extended, high resolution mapping. Our system is designed to maximize the resolution possible for such volumetric methods, while working on an unbounded space. 
\end{abstract}
\section{INTRODUCTION}

Simultaneous Localization and Mapping (SLAM), is an important task for robotics for applications such as navigation, path planning, and obstacle avoidance. A robot that performs SLAM should be able to create a map of an unknown environment and localize itself within the map as it traverses. The importance of the problem has led to extensive research over the past few decades.

With advances in RGB-D cameras, visual SLAM methods have improved to construct dense RGB-D maps in real time using GPU hardware, leading to advances in not only mapping but more generally in robotic perception. A common method is KinectFusion \cite{Newcombe11ismar}, which popularized dense reconstruction for a predefined volume in real time. Other methods \cite{Whelan13iros, Roth12bmvc} have modified the KinectFusion algorithm to be spatially extended for mapping larger, building-scale environments instead of a predefined volume. Such methods allow for millimeter level reconstructions of the world. 

Dense mapping has created a data domain for uses beyond traditional SLAM and allow for applications such as object discovery and detection \cite{finman13ecmr, karpathy13icra}, as well as motion planning \cite{Wagner13icra}. Such applications would still benefit from higher density representations, but the resolution of current methods is limited by the capacity of RAM on the GPU.

\begin{figure}
\begin{center}
\includegraphics[width=0.8\columnwidth]{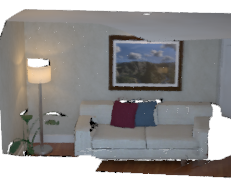}
\includegraphics[width=0.8\columnwidth]{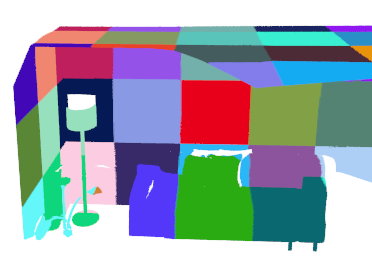}
\end{center}
\caption{\textbf{Top:} A colored model of the ICL-NUIM Livingroom dataset \cite{handa14icra}. \textbf{Bottom:} A reconstruction of the same scene with segments showing regions mapped by a volume. Each volume has the maximum resolution that can be loaded into GPU RAM. Volumes are dynamically placed so they are only computed where from the camera is. }
\label{fig:many_cubes}
\end{figure}

In this work, we propose a method allows for one to maximize the resolution of models obtainable by volumetric method such as those in \cite{Curless96siggraph}. We provide for the ability to increase resolution of different areas, depending on the requirements of the application. We build on the KinectFusion method by placing multiple small volumes that, due to their reduced physical dimensions, have higher resolution over a smaller patch of the world (See Fig. \ref{fig:many_cubes}). Each smaller volume is passed in and out of the GPU memory as needed for mapping. Such a method can produce identical results to KinectFusion with unbounded GPU RAM.  We then propose a dynamic volume allocation method that both provides higher resolution where it is needed, based on user defined parameters and allows for spatially extended regions. By adding new volumes and removing old ones, the mapping space can be dynamically changed as the camera moves through the world.

In this paper, we present two primary contributions:

\begin{enumerate}
\item A novel high resolution mapping system using multiple mapping volumes to achieve maximally dense reconstructions.
\item A volume placement method for spatially extended mapping and dynamic resolution modifications based on the environment.
\end{enumerate}
We will also release an open source implementation built off of PCL Kinfu \cite{Rusu11icra}.

The paper is structured as follows. First, we discuss related work in the field and how our approach differs. Second, we detail how our multi-volume high resolution mapping method works. Third, we describe our dynamic volume allocation method. Last, we evaluate our work against the state-of-the-art and show how our method has similar accuracy with higher resolution.

\section{Related Work}

Dense reconstruction methods have become popular since RGB-D sensors have become ubiquitous. While visual SLAM systems were developed before RGB-D sensors \cite{comport07icra} and there is much work in monocular SLAM \cite{Engel14eccv, Newcombe11iccv, Klein07ismar}, we will cover only the more recent dense RGB-D SLAM methods here.

There has been significant literature in RGB-D Mapping \cite{Zeng12cvm,Zhou13Siggraph,Zhou13iccv,Canelhas13iros,Pirker11bmvc,Stueckler13jvcir}. Henry \emph{et. al} \cite{henry12ijrr} introduced one of the first RGB-D mapping algorithms. Their method makes a surfel map of the environment by using visual features and geometric tracking to construct a pose graph. Feature-based methods continue to be a popular approach \cite{endres12icra}, though many feature-based methods do not take full advantage of the entirety of the data, which can produce improved results in reconstruction \cite {Newcombe11iccv}.

There have been several keyframe-based approaches for RGB-D mapping. Tykkala \emph{et. al} \cite {Tykkala13iros} fuses new camera data into an existing keyframe model until a new pose is added. Meilland \emph{et. al} \cite{Meilland13iros} presents an impressive system building on their previous work \cite{Meilland13iccv, Meilland13icra} that combines both the benefits of keyframe and volumetric-based mapping representations. Their method works for large-scale environments in a memory efficient manner. In order to represent the 3D features of a small cluttered scene many small keyframes are required, which could potentially cause tracking drift or failure.

There have been many extensions to KinectFusion, primarily allowing for extending the mapping region beyond a static volume \cite{Roth12bmvc, Whelan13iros}. Whelan \emph{et. al} presents a similar method to our own which extends KinectFusion by moving the TSDF volume through space by utilizing a circular buffer on the GPU. This extends the high resolution capability to building-scale environments. However, this method is still resolution constrained by GPU RAM. Other work on dense RGB-D camera tracking was done by Steinbruecker \emph{et. al} \cite{Steinbruecker13iccv}, estimating a warping function between an image's geometric and photometric information and building an octree-based TSDF on a GPU. By using an octree representation building on \cite{Wurm10icraw}, the system can easily handle multiple resolutions and can be used in conjunction with our work. These methods exploit both the capabilities for the CPU and the GPU to balance the computation load effectively for real time mapping. Chen \emph{et. al} \cite{chen13tog} use a hierarchical structure an efficient memory passing between GPU and CPU to minimize unnecessary computation, but requires domain knowledge to determine the correct hierarchical parameters. Nie{\ss}ner \emph{et. al} \cite{nieBner13tog} present a powerful voxel hashing method for large speed up. However they do not support dynamic resolution modification, and through hashing potentially remove data and thus do not guarantee optimal point-wise estimation.

Henry \emph{et. al} \cite{Henry133dim} uses small patch volumes to segment a mapped space into manageable chunks. Each patch volume is modeled as an independent signed distance function (SDF) and the patches are used for handling loop closures. Having multiple SDFs is similar to our own method, but they use the SDFs for consistent loop closures after they have mapped a scene and not for mapping. 

\begin{figure}
\begin{center}
\includegraphics[width=0.4\columnwidth]{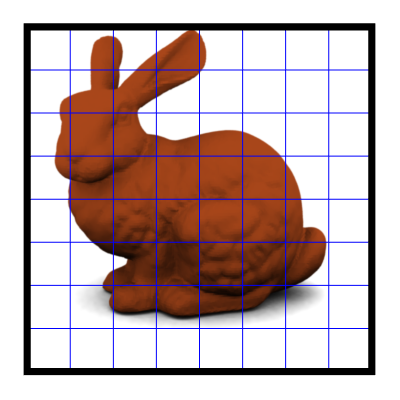}
\includegraphics[width=0.4\columnwidth]{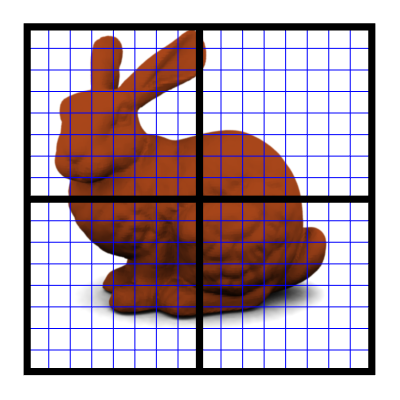}
\end{center}
\caption{\textbf{Left:} 2D representation of KinectFusion volume with 8 linear voxels scanning a bunny. \textbf{Right:} Four volumes each with eight linear voxels scanning a bunny of the same size. In this case we see that we double the linear resolution by using four volumes each with half the length.}
\label{fig:highlow2}
\end{figure}

\section{High Resolution Mapping}

We are trying to solve the problem of using a series of depth maps to reconstruct the maximum resolution 3D model that generates those depth maps. As all points read by the camera are noisy, this can be viewed as attempting to find the center of the probability distribution over measurements for each point, while maximizing the number of points modeled. Here we present the problem more precisely based on the input and output relationship, metrics.

\subsection{Input and Output}

Our algorithm is designed to output the 3D reconstruction of a fixed size cubic region of physical space using a sequence of provided depth maps. The input to the algorithm is therefore as follows:

\begin{itemize}

\item The linear dimension, $l \in \mathbb{R}^+$, of the physical region to be scanned. Our region is assumed to be a cube, so this is the side length of a cube to be scanned.

\item The linear number of voxels, $r \in \mathbb{N}$, of the model used during 3D reconstruction. Our region is assumed to be a cube, so this is the side length of a virtual cube of voxels.

\item A series of $i$ $n \times m$ depth frames, $D^i \in \mathbb{R}^{+nm}$ provided by a camera. We assume each frame is generated from a camera by a noisy, physically realizable model.

\item The camera intrinsic matrix, $K \in \mathbb{R}^{3\times3}$. The camera intrinsics provide information about the mapping from the values in a depth frame to the physical location of the surface that generated that reading \cite{Hartley03book}.

\end{itemize}

During the scanning process a collection of voxels is stored as a cube of voxels. Each voxel stores the signed distance from each voxel to the nearest surface, which is then truncated if this distance is too large. The collection of voxels is known as the truncated signed distance function (TSDF) \cite{Curless96siggraph} volume. The number of voxels $r^3$ in the TSDF volume acts as a metric for the resolution of the reconstruction (see \ref{subsec:metrics}). 

 The depth frames are each a 2D matrix, with each value representing the depth to the nearest surface of a given ray. The camera intrinsic matrix is used to determine the angle of a given ray described by the depth map.

Each measurement in $D^i$ is itself noisy with an unknown noise function. The core function of this algorithm is to integrate all measurements in $D^i$ to determine the true generating model.

The algorithm outputs the following:

\begin{itemize}

\item A set $V$ of vertices, consisting of elements $v^j \in \mathbb{R}^3$ for $j = [1, |V|]$. Each vertex is the physical location of a 3D point relative to the camera initialization location.

\item A set $N$ of normals, consisting of elements $n^j \in \mathbb{R}^3$ for $j = [1, |N|]$.  $n^j$ corresponds to the normal for vertex $v^j$.
\end{itemize}
We note that the magnitude of $V$ is upper bounded by $r^3$. That is, the number of vertices in the reconstruction cannot exceed the number of voxels in the model used for reconstruction.

\subsection{Metrics}
\label{subsec:metrics}

In the ideal case, each vertex represents the location of a surface, with location identical to that of a point on some surface relative to the initial camera location. To capture the error in these distance, we define the mean distance from each vertex to the nearest true surface as the \textit{accuracy} of the model. As a second constraint, for each voxel in the reconstructive model, if a surface crosses through that voxel, then a vertex is created for that voxel. Increasing the number of vertices in the output will be defined as increasing the \textit{resolution} of the model. 

%
%
%
%
%
%
%
%

\section{Reconstruction Algorithm}

Those who are familiar with the KinectFusion algorithm will notice that the steps listed below follow a similar structure. The key modification with our algorithm is that it allows for the ability to increase the resolution of the reconstruction to a user specified degree in exchange for computation time.

\begin{figure}
\begin{center}
\includegraphics[width=0.7\columnwidth]{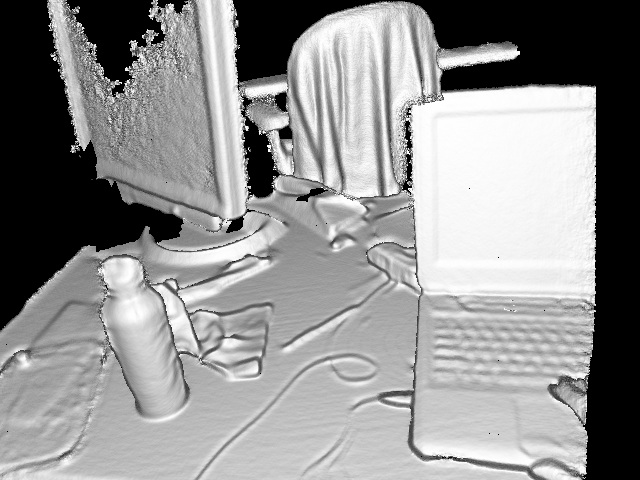}
\end{center}
\caption{Image generated from 4 1.5 meter TSDF subvolumes.}
\label{fig:highlow}
\end{figure}

\subsection{Initialization}

This algorithm allows for the number of voxels $r^3$ to be a user tunable parameter. However, future steps in our algorithm rely on being able to perform computation on the TSDF volume on a GPU. In the case where the user-provided $r$ results in too many voxels for the entire volume to be stored in GPU memory, the TSDF volume will be broken into a set of TSDF subvolumes $H$. Each subvolume $h\in H$ acts as a TSDF volume and is used in reconstruction, but are separated so that computation can be performed on them independently. 

We define $r_\text{GPU}$ to be the maximum number of linear voxels that can fit into GPU memory. The linear number of TSDF subvolumes initialized is $\frac{r}{r_\text{GPU}}$.

The initial location of the camera is initialized as $(0,0,0)$. The back, center voxel of the TSDF volume is additionally initialized to this location. In order to achieve this, each subvolume is placed such that they form a cube with back center at $(0,0,0)$. Each subvolume stores the translation $h_t$ of its back center voxel relative to $(0,0,0)$, its linear number of voxels $h_r$ and its linear physical dimension $h_l$. For reasons mentioned Section \ref{subsec:raycasting}, each subvolume overlaps by two pixels, but this will be ignored for the majority of the analysis, as it only results in a constant factor decrease in the total volume size. 

\subsection{Depth Map Conversion}

For each frame $i$, the depth map $D^i$ is converted into a collection of vertices. A given vertex in coordinates of the current frame $v^i_l(x,y) = K^{-1}[x, y, 1]^TD^i(x,y)$ for pixel coordinates $x = [0, m-1]$ and $y = [0,n-1]$. This is done on the GPU on a per pixel basis. The normals are then computed for each vertex: $n_l^i(x,y) = (v^i_l(x+1,y) - v^i_l(x,y))\times(v^i_l(x,y+1) - v^i_l(x,y))$. 

Using the transformation matrix derived from the previous camera tracking iteration (see Section \ref{subsec:tracking}), the global position of each point is then derived. The global transformation matrix $T^{i-1} = [R^{i-1}|t^{i-1}]$, with translation vector $t^{i-1} \in \mathbb{R}^3$ and rotation matrix $R^{i-1} \in SO(3)$. The global vertex location is then $v_i(x,y) = t^{i-1}v^i_l(x,y)$ and global normal is $n_i(x,y) = R^{i-1}n_l^i(x,y)$. This puts our vertex in normals in the global coordinate frame defined, which is necessary for future steps.

\subsection{Camera Tracking}
\label{subsec:tracking}

The transformation matrix between the vertices derived from the previous raycasting step and the current depth map conversion step is determined. This is run by performing a GPU based iterative closest point (ICP) algorithm, the details of which can be found in \cite{Newcombe11ismar}. 
The transformation is then applied to $T^{i-1}$ to determine the current global transformation matrix $T^i$.

\subsection{Memory Passing}

\tikzstyle{start} = [rectangle, rounded corners,minimum width=.5cm, minimum height=.7cm, text centered, text width=2cm, draw=black, fill=red!30]
\tikzstyle{frame} = [rectangle,rounded corners, minimum width=.5cm, minimum height=.7cm, text centered, text width=2cm, draw=black, fill=blue!30]
\tikzstyle{volume} = [rectangle, rounded corners,minimum width=.5cm, minimum height=.7cm, text centered, text width=2cm, draw=black, fill=green!30]
\tikzstyle{arrow} = [thick,->,>=stealth]
\tikzstyle{line} = [draw, -latex']
\tikzstyle{every node}=[font=\small]
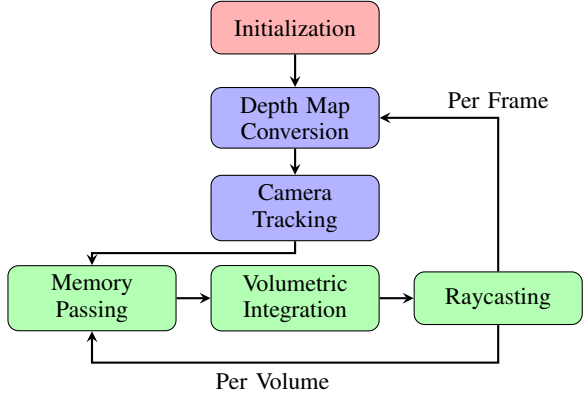
\begin{figure}
\begin{center}
\begin{tikzpicture}[node distance=2.7cm]
\node (start) [start, xshift=2.7cm] {Initialization};
\node (dmc) [frame, below of=start, yshift=1.5cm] {Depth Map Conversion};
\node (ct) [frame, below of=dmc, yshift=1.5cm] {Camera Tracking};
\node (mm) [volume, below of=ct, yshift=1.5cm, xshift=-2.7cm] {Memory Passing}; 
\node (vi) [volume, right of=mm] {Volumetric Integration};
\node (rc) [volume, right of=vi] {Raycasting};
\draw [arrow] (start) -- (dmc);
\draw [arrow] (dmc) -- (ct);
\draw [arrow] (ct) to[|-|] (mm);
\draw [arrow] (mm) -- (vi);
\draw [arrow] (vi) -- (rc);
\draw [arrow] (rc.south) |- ++(-3, -.5) node[ below] {Per Volume} -| (mm);
\draw [arrow] (rc.north) |- node[above] {Per Frame} (dmc);
\end{tikzpicture}
\end{center}
\caption[Flowchart for model creation.]{Flowchart for model creation. At the beginning of each dataset, TSDF volumes are initialized. For each frame, the frames depth map is converted to global coordinates and the camera transformation from the previous frame to the current one. A volume is passed to GPU memory, where volumetric integration and raycasting are then performed to update our TSDF volumes and obtain the raycasted depth map. \textbf{Red:} Performed once per dataset. \textbf{Blue:} Performed once per frame. \textbf{Green:} Performed on each volume for each frame. }
\label{progression}
\end{figure}

To store the amount of data required to have multiple TSDF subvolumes, they generally need to be stored in CPU memory, as we constructed the subvolumes such that only one may fit in GPU memory. Generally a computer has many times more CPU memory than GPU memory. Because of this, and the benefits of virtual memory, we can store more volumes in CPU memory than in GPU memory (see section VI.B.). Prior to a volumetric integration step, one TSDF subvolume is uploaded to the GPU for point volumetric integration and raycasting. After raycasting, this volume is downloaded to CPU memory. This happens for each subvolume, upon which the final raycasting step is complete for that frame.

\subsection{Volumetric Integration}
During this step we update our TSDF subvolumes, which store our beliefs of which surfaces are relative to our global origin. Each subvolume $h$ stores a set of values $h_d^{i}(t_l)$ where $t_l = (t_l^x, t_l^y, t_l^z) \in \mathbb{R}^3$ represents the voxel coordinate, and $h_d^{i} \in \mathbb{R}^{r^3}$ stores the believed distances to the nearest surface for each voxel in subvolume $h$ on the $i$th frame.

Each GPU thread is assigned an $(t_l^x, t_l^y)$ location, for $t_l^x = [0, r-1]$ and $t_l^y = [0, r-1]$. Each thread then iterates over each $t_l^z = [0,r-1]$ for its given $(t_l^x, t_l^y)$. Each voxel's distance to the nearest surface  is determined using its $t_l$ coordinate. However because each subvolume is translated relative to the origin by translation $h_t$, the coordinate to use to calculate distance is then $t_s = t_l + h_t$. From here the physical model distance is calculated as $d_m = \frac{l}{r} ||t_s||$, which represents the distance from the camera origin to the voxel.

In order to determine the measured distance for that frame, each $t_s$ voxel position is then back projected to pixel coordinates to get an $(x,y)$ location. Using this the distance to the nearest surface is calculated as $dist = D^i(x,y) - d_m$. $dist$ is then constrained to be within some truncation bounds proportional to the physical distance of a voxel in order to prevent unnecessary computation. From here we can calculate the updated TSDF value $h_d^{i}(t_l) = \frac{w^{i-1}h_d^{i-1}(t_l) + w^i dist}{w^{i-1} + w^i}$, for each voxel, which is the belief of the distance from that voxel to the nearest surface. $w^i$ is a linearly increasing weighting function, up to some maximum.

\subsection{Raycasting}
\label{subsec:raycasting}
During this step we derive the depth map constructed from our TSDF volume. As mentioned above, the raycasting step is run once separately per subvolume, each after the matching volumetric integration step.

Prior to the memory passing step a pixel distance map $v_r \in \mathbb{R}^{+}$ and normal map $n_r \in  \mathbb{R}^{+3}$, each of size $m \times n$ are initialized. Based on the camera intrinsics, a ray can be cast from each pixel location $(x,y)$ in the camera space through the TSDF volume. As each ray iterates through the TSDF volume space, checks for the value of the TSDF voxel it is currently in. If the ray passes through a change of sign between voxels, it is noted to have crossed through a surface. By trilinearly interpolating over the voxels on either side of the zero-crossing, the location of the surface is determined. Because the side of each voxel is known, the raycasted voxel location $v_r(x,y)$ is determined. Similarly the derivative of the TSDF as the zero-crossing is used to determine the normal value $n_r(x,y)$  for this pixel.

However because we are only working on a given subvolume, a few modifications need to be made. While raycasting, calculation is only done while the ray is within the TSDF subvolume, or until it finds a zero-crossing. As a ray is cast from a camera location, its voxel position in global space can be described as a position $t_r \in \mathbb{R}^3$. The voxel position in volume coordinates is then $t_r' = t_r - h_t$. We then read values of the voxels at $t_r'$ when searching for zero crossing. When determining the global physical locations $g$ of each voxel, the translations must be added back to our new indices, to accurately reflect the physical distance the ray travels to our translated volume. This gives us $g = h_l*(t_r' + h_t)$. When a zero crossing is found, the value is trilinearly interpolated over neighboring voxels to determine $v_r(x,y)$, and the derivative of the TSDF at the zero crossing is calculated to determine $v_n(x,y)$.

It is worth noting that all raycasting steps modify the same map. This means each pixel coordinate is searched over by each subvolume, even though all edit the same map. In order to guarantee that we use the correct value in our voxel map, we take the value with the shortest raycast distance from the local camera location. This guarantees that we always take the value of the surface visible by the current camera location. To trilinearly interpolate successfully each subvolume must overlap by 2 voxels.

\subsection{Comparison to KinectFusion}

We show that when creating a cube of multiple volumes the results are identical to that of running the KinectFusion algorithm on one volume with the same total number of voxels. Because the volumetric integration and raycasting steps are the only ones modified, if we can show they give identical outputs to KinectFusion, then the entire algorithm also does so. For sake of simplicity we will ignore the overlapping voxels in our volumes when discussing the number of voxels, which is accounted for by requiring that each volume be larger by the size of 2 voxels.

During the volumetric integration step, each voxel is independent of each other voxel. That is, the calculation performed only depends on the TSDF value of the given voxel, and the depth of the ray back projected from that voxel. All modifications done by our algorithm guarantee that the voxel is associated with the correct physical location, but do not affect the TSDF value or the depth value. Due to the fact that all computation depends on each voxel independently, that voxels exist in separate volumes does not affect the correctness of the computation.

In the raycasting step, unlike the volumetric integration step, the voxels are not independent. Due to trilinear interpolation, each voxel depends on each neighboring voxel. However given that the volumetric integration step is identical to KinectFusion, upon overlapping the volumes, the overlapping voxels will have identical values. Therefore during the trilinear interpolation step, every voxel for which trilinear interpolation is possible will have a set of neighbors identical to that of KinectFusion. By then only using the shortest distance zero-crossing found by each ray, the depth and normal maps are identical to those in KinectFusion.

\subsection{Benefits}
\label{sec:benefits}

This algorithm is designed explicitly to create high-resolution volumes of a fixed area. By using the TSDF volume techniques of KinectFusion we create a more accurate 3D model than any individual frame. Then by allowing for a number of subvolumes, and thereby voxels, only bounded by disk space, we allow for resolution functionally bounded only by the noise in the camera which generates the depth frames. This is compared to the resolution bounded by GPU size in KinectFusion. 

\section{Dynamic Volume Allocation}
Our goal is again to create a 3D reconstruction of a noisy measurement model while optimizing over accuracy and resolution. However in this case we present the additional requirement that resolution of the model in a given region be variable based on user input.

\subsection{Input and Output}

The output relationship of dynamic allocation is identical to that of the high-resolution reconstruction.

During runtime the algorithm will keep a list $H$ of active subvolumes it is processing on. Each subvolume $h \in H$ is initialized by 3 parameters: $h_l$, the linear physical dimension, $h_r$, the linear number of voxels, and $h_t$, the translation relative to the camera origin. A volume can be added to the set at any point during execution, and will be processed in subsequent frames. To remove a volume from processing, it is removed from the set of volumes. The point cloud for the removed volume can then be downloaded to main memory. Thus the additional input parameters are:

\begin{itemize}
\item 
$f(\cdot)$, which can take in any data available during a given frame. It returns a list of volume parameters $(h_p^1,...,h_p^n)$ for the volumes to be added, where $h_p = (h_l, h_r, h_t)$ for some subvolume.

\item 
$g(h, \cdot)$, which is run for all $h \in H$. The function removes the volume from $H$ if the conditions it specifies are met. It can additionally take in any other parameters available in the system.
\end{itemize}

By defining $f$ and $g$, this algorithm can be extended to a multitude of use cases. For example, the work of Kaparthy \emph{et al.} \cite{karpathy13icra} could be applied to create higher resolution models for areas with object-like properties. 

%
%
%
%
%
%
%


\begin{figure}
\begin{center}
\includegraphics[width=0.4\columnwidth, trim=120mm 30mm 120mm 30mm, clip]{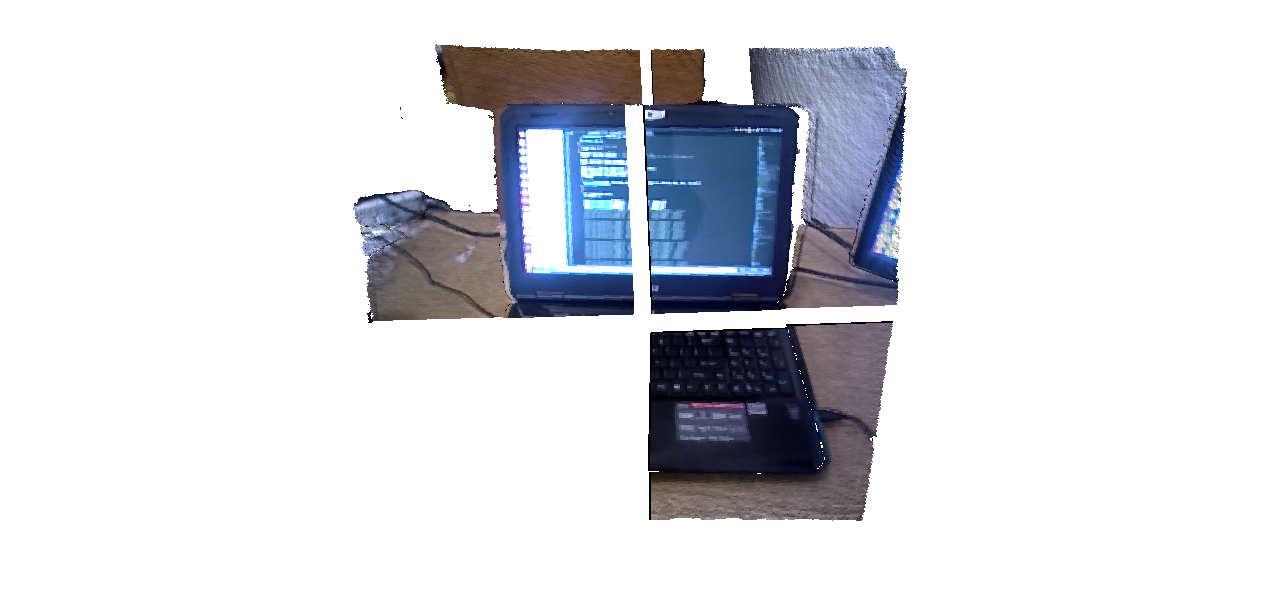}
\end{center}
\caption[Example of spaced out TSDF subvolumes.]{Volumes can be placed anywhere relative to the initial camera location. Here we scan with three spaced out volumes, and choose not to allocate a volume in the bottom left corner for demonstration purposes.}
\label{spaced}
\end{figure}
\setlength{\textfloatsep}{4pt}
 \begin{algorithm}
 \caption{The update steps performed for each volume.}
 \label{algorithm:dynamic}
 \begin{algorithmic}[1]
 \Procedure{Volume Update($h$, $D$, $v_r$, $n_r$)}{}
 \State $uploadToGPU(h)$
\For{$(x,y)$ in $([0,r-1],[0,r-1])$}
\For{$z$ in $[0,r-1]$}
\State $t_s \leftarrow t_l + h_t$
\State $(x,y) \leftarrow backProject(t_s)$
\State $dist \leftarrow D^i(x,y) - \frac{l}{r} ||t_s||$
\State $h_d(t_l)^i \leftarrow updateWeight(h, x, y, dist)$
\EndFor
\EndFor
\For {$(x,y)$ in $([0,r-1],[0,r-1])$}
\State $t_r \leftarrow castRay(x,y,(0,0,0)) - h_t$
\State $t_r' \leftarrow t_r - h_t$
\While{$inVolume(t_r - h_t)$}
\State $t_r \leftarrow castRay(x,y, t_r) - h_t$
\State $t_r' \leftarrow t_r - h_t$
\If{$zeroCrossing(t_r')$}
\State $g \leftarrow h_l*(t_r' + h_t)$
\State $v_r(x,y) \leftarrow calculateVertex(g)$
\State $v_n(x,y) \leftarrow calculateNormal(g)$
\EndIf
\EndWhile
\EndFor

\State $uploadToCPU(h)$
 \EndProcedure
  \end{algorithmic}
 \end{algorithm}

\subsection{Insertion and Removal}

After the final raycasting step an additional insertion and removal step is added. During this step the removal algorithm $f(\cdot)$ is first run. Each subvolume to be added is specified with $h_l$, the linear physical dimension, $h_r$, the linear number of voxels, and $h_t$, the translation relative to the camera original. Each subvolume specified by $f(\cdot)$ is then marked for addition.

$g(h, \cdot)$ is then run for each $h \in H$. Note that this algorithm can use any parameter of the algorithm, including the set of subvolumes marked for addition.  If a subvolume specified has identical parameters to one marked for addition, that volume instead will not be marked for addition. This provides a mechanism for $f(\cdot)$ and $g(h, \cdot)$ to interact. Each subvolume specified is then marked for addition.

Because each TSDF volume can be described as a point cloud, the volumes marked for deletion have their point cloud representations downloaded to CPU memory. The volumes marked for addition are then added to $H$, and used during the volumetric integration and raycasting steps of future iterations.

\subsection{Dynamic Mapping for SLAM} 
To perform large-scale mapping we propose a tool for dynamically placing subvolumes in areas expected to have a high density of surface vertices. There are three tunable parameters in this system: The linear number of voxels in each TSDF subvolume $d_v$, the linear physical dimension in each subvolume $d_l$, and the maximum number of subvolumes to exist at once time $n$. Our system dynamically places $n$ subvolumes $h$ each with $h_l = d_l$ and $h_v = d_v$ in locations based on the following process.

We define the origin of the global coordinate frame as the initial location of the camera. Assuming voxels of fixed size, we can view this coordinate system as having length units in voxels rather than a physical length. We allow subvolumes to be placed on this coordinate system such that their translations from the origin are constrained to $h_t = k*(d_v-2)$, where $k \in \mathbb{Z}^3$. This allows the subvolumes to be placed on a 3D grid, where each cell is the size of a subvolume. The $(-2)$ term is so that adjacent cubes do not create a gap in the raycasting result and will be excluded for simplicity of notation. We do not allow multiple subvolumes to have the same translation, so a subvolume's placement in this grid defines a unique property of the subvolume.

During the raycasting operation, for each pixel $p = D^i(x,y)$ in the depth frame, we calculate the the endpoint $e = (e^x, e^y, e^z) \in \mathbb{R}^3$ of the ray generated from $p$ based on the depth $r_d$ and direction $r_v = (r_v^x, r_v^y, r_v^z) \in \mathbb{R}^3$ from $p$. Using this we calculate the location in global coordinates of where that depth value maps. 
$$e^x = \frac{r_v^x}{\sqrt{(\frac{r_v^z}{r_v^x})^2 + (\frac{r_v^y}{r_v^x})^2 + 1}}$$
$$e^y = \frac{r_v^y}{r_v^x}*x, e^z = \frac{r_v^z}{r_v^x}*x$$

We then convert each endpoint to a coordinate $l$ which is constrained to $h_t = k*(d_v)$ as the subvolume locations are $l = \frac{(e^x, e^y, e^z)}{d_l} - (\frac{(e^x, e^y, e^z)}{d_l} \mbox{ mod } d_v)$

Any ray with endpoint $l$ maps uniquely to a subvolume. We define $c(l)$ as the number of rays that end at the subvolume mapped to $l$. The set $L$ is defined as all subvolume locations with non-zero $c(l)$. Our insertion function takes in the current subvolume set $H$ and all $c(l)$ for $l \in L$, so the function signature is $f(H, L, c(L))$. We then only keep the subvolumes with the $N$ largest $c(l)$ values. Our removal function $g(h, \cdot)$ removes a subvolume which has been marked as having falling out of the top $n$ subvolumes. In order to prevent volumes from switching in and out too often, we also set a threshold that a subvolume to be added needs to have a count larger than the lowest count subvolume by some constant factor.

\begin{figure}
\begin{center}
\includegraphics[width=.6\columnwidth]{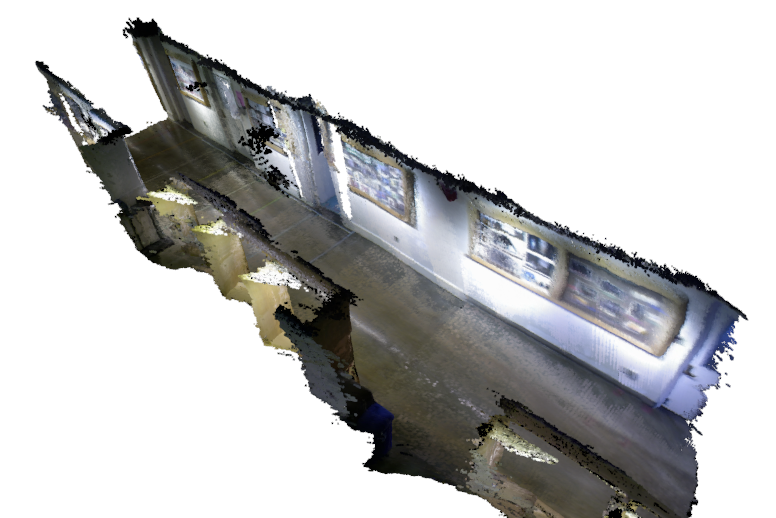}
\includegraphics[width=.35\columnwidth]{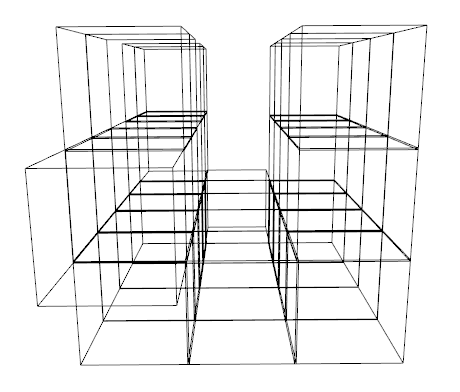}
\end{center}
\caption[Scan of a hallway with cube placement.]{\textbf{Left:} Scan of hallway. This scan was taken while simply walking forward, yet still captures both walls and the floor at high resolution. 2m subvolume were dynamically allocated such that at most 16 were live at one time. This image consists of 24 subvolumes. \textbf{Right:} Volume placement on a short hallway scene. We see no subvolumes allocated in the region between the walls.}
\label{fig:hallway}
\end{figure}

\section{Experiments}

The key benefit of our algorithm is the ability to increase the resolution of our model without sacrificing accuracy, but at the cost of computation time. Therefore to quantitatively evaluate our algorithm we use two metrics:

\begin{itemize}
\item
Runtime performance of the algorithm as a function of the number of TSDF volumes being used.

\item
Comparison against the PCL implementation of KinectFusion (KinFu) \cite{Rusu11icra} using benchmarks developed by  \cite{handa14icra}.
\end{itemize}

We further discuss qualitative results of our algorithm.

\section{Hardware}
Our computer uses an nVidia GeForce GTX 880M graphics card with 8GB GPU memory, along with a 2.7GHz Intel Core i7 4800MQ processor. It has 32GB main memory, along with a 1TB SSD. We allocated an additional 64GB of swap space in order to guarantee the function of our algorithm. We used an ASUS Xtion Pro camera for all experiments. 

\begin{table}
\caption{Model to Ground-Truth Error}
\centering
\begin{tabular}{|llll|}\hline
& traj0 & traj1 & traj2 \\ \hline
KinFu Volume & & & \\
vertices(m) & 1497300 & 1465488 & 506229 \\
mean(m) & 0.0309 & 0.0534 & 0.0213 \\
median(m) & 0.0215 & 0.0375 & 0.0134 \\
std(m) & 0.0282 & 0.0542 & 0.0254 \\ \hline
Eight Dense Volumes & & & \\
vertices & 7318191 & 7354923 & 2192196 \\
mean(m) & 0.0335 & 0.0537 & 0.0222 \\
median(m) & 0.0253 & 0.0368 & 0.0152 \\
std(m) & 0.0315 & 0.0545 & 0.0250\\ \hline
\end{tabular}
\end{table}

\section{Runtime Performance}
The fundamental tradeoff of our algorithm is speed versus resolution.  As the number of volumes we use increases we expect the time transferring data from GPU memory to CPU memory to account for the majority of the processing time. Because this happens for every volume on each frame, we expect the time to be linear with the number of volumes, until the computer switches to swap memory.

We ran the following experiment to validate our runtime expectations. On a pre-recorded 100 frame real world dataset we dynamically allocated a fixed number of volumes with 512 linear voxels on the first frame. For 2 to 7 volumes we used 1.5m volumes. In order to guarantee there existed unique data for each volume, we decreased the size of each volumes as the number of volumes increased.

Our results show the seconds per frame is linear with the number of volumes, with each volume adding approximately .17s per frame. By running the algorithm with 48 volumes we demonstrate the ability to generate a 48x increase in resolution compared to a single volume. We also found that both PCL KinFu and our implementation ran at approximately 11 frames per second for one volume. It is worth noting that this is less than the expected 30 frames per second because the images were read out of memory instead of a live camera feed. At 64 volumes the computer beings using swap memory, and the time increases to approximately 150 seconds per frame.

%
%

\section{Comparison with Baseline}

We compare our algorithm to PCL KinFu on the synthetic dataset developed by Handa \emph{et al.} \cite{handa14icra}. This dataset provides a series of RGB and Depth frames that are structured simulate the camera movement through a virtual living room. Also provided is an .obj model for each dataset as ground truth. In order to accurately simulate a depth camera, Handa \emph{et al.} added noise as described in \cite{Barron13cvpr} to each frame.

For our experiments we limited the range of camera scenes to those contained by a 6m KinectFusion volume. For the experiments we ran each mapping algorithm on a scene in order to generate a 3D map. We then use the open source Cloud Compare\footnote{http://www.danielgm.net/cc/} to align the reconstruction to the ground truth model. This provides the translation of each point in the reconstruction relative to the nearest mesh triangle in the ground truth model. Using the statistics provided by \cite{handa14icra}, we then determine the mean, median, and standard deviation for each reconstruction's translation, the results of which can be seen in Table 1. For traj0 and traj1 we ran PCL KinFu with a 6m volume size, and our algorithm with a set of 8 volumes forming a cube of 6m (3m per volume). For traj2 we we ran PCL KinFu with a 3m volume size, and our algorithm with a set of 8 volumes forming a cube of 3m (so 1.5m per volume). All trajectories used 512 linear voxels per volume.

Table 1 shows that our algorithm generates on average 4.75 times as many vertices, indicating a 4.75 time increase in resolution by our definition. Reconstruction accuracy is in line with KinFu, with the average mean error increased by 4.2\%. Due to the relatively small number of points used to generate the synthetic meshes, we do not expect to see an improvement in model reconstruction statistics relative to KinectFusion, even for large volumes, due to the fact the synthetic dataset is almost entirely flat surfaces, reconstruction resolution has a minor impact on error. Given the small difference in quality and high number of vertices increased, we believe this highlights the algorithm's ability to generate high-resolution datasets without reducing accuracy. As a matter of future work it would be useful to generate a dense synthetic dataset and determine if our model reduces mesh reconstruction error.

\begin{figure}
\begin{center}
\includegraphics[width=0.8\columnwidth, trim=150mm 0mm 130mm 60mm, clip]{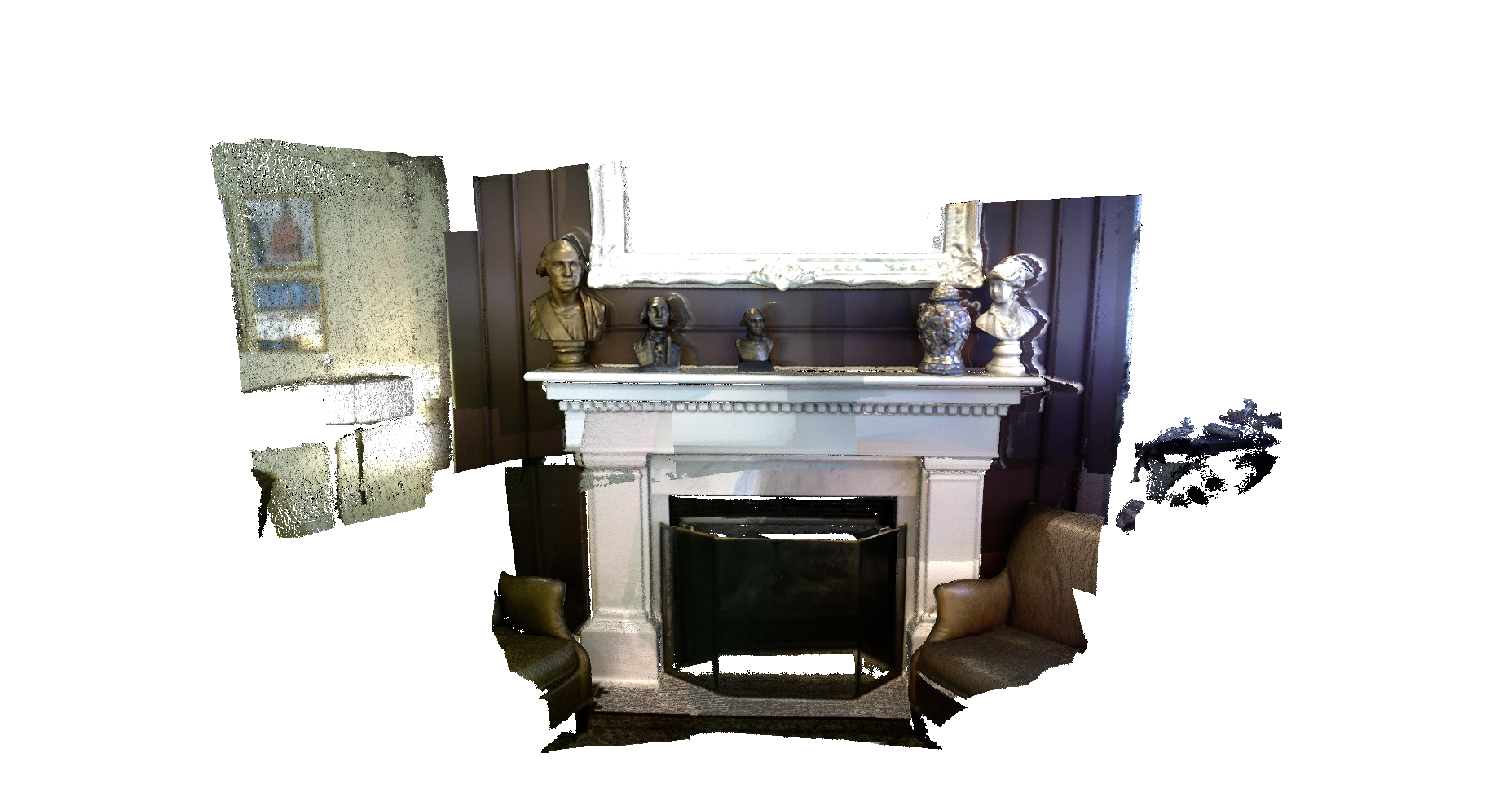}
\includegraphics[width=0.8\columnwidth, trim=120mm 0mm 100mm 0mm, clip]{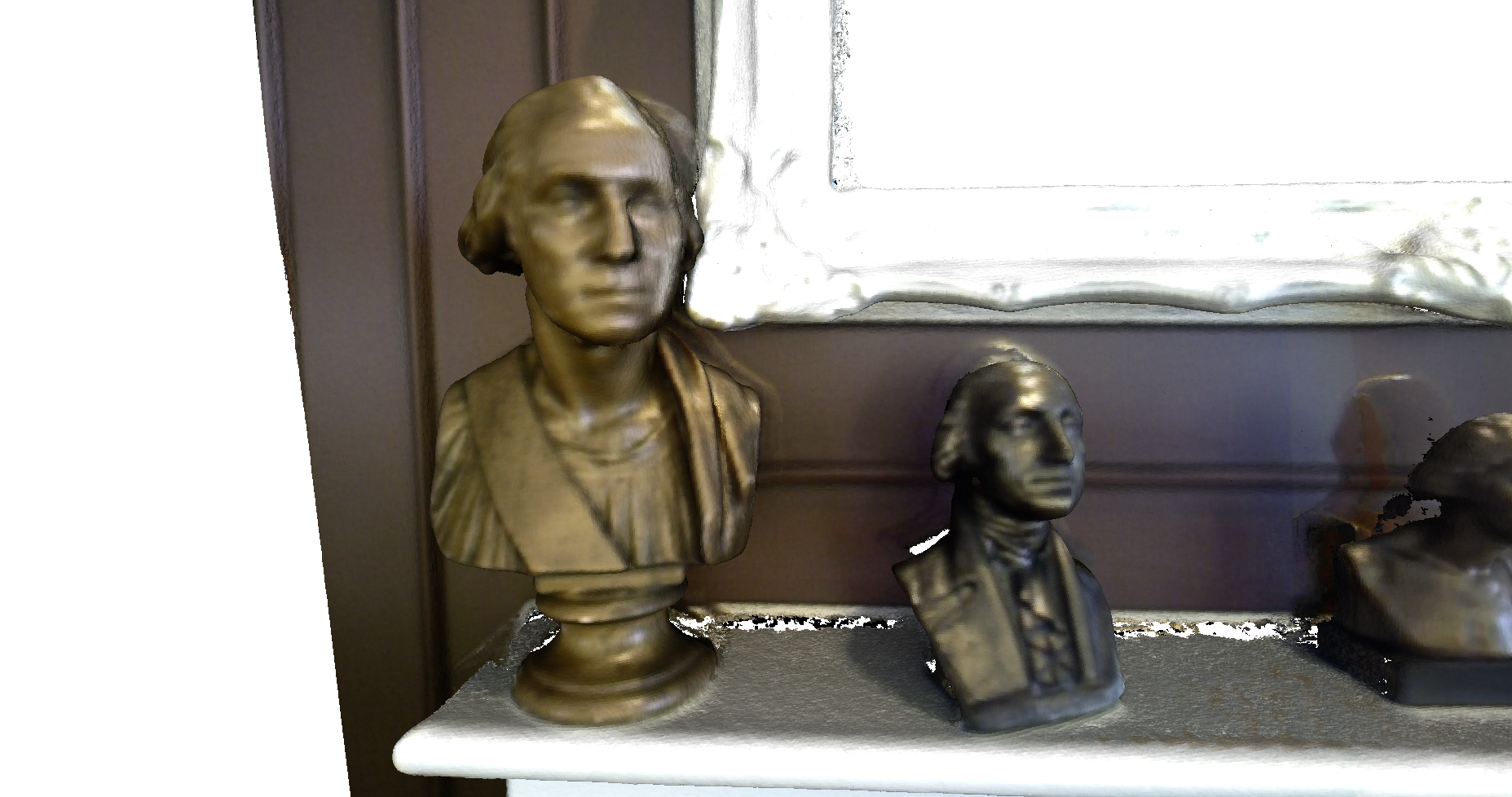}
\end{center}
\caption[Colored fireplace and mantel scan at Kendall Hotel.]{\textbf{Top:} Fireplace scene at the Kendall Hotel in Cambridge, MA. \textbf{Bottom:} Zoom of two busts from the same model. Model was generated using dynamically placed .75m volumes. The color misalignment is due to a misalignment in the colors with the 3D model, and is not an issue with the 3D model itself.}
\label{fig:fireplace}
\end{figure}

\section{Qualitative Results}

We tested our dynamic volume allocation with 3m volumes while walking down a hallway, as seen in Figure \ref{fig:hallway}. To obtain this dataset the camera was not rotated as it was translated through the hallway, highlighting the ability to obtain large scale datasets easily. We found that large datasets like this often had drift due to accumulated error in ICP, which can be resolved by future work improving visual odometry.

The fireplace seen in Figure \ref{fig:fireplace} highlights power of our algorithm. The figure was generated by a 40 second scan of the fireplace and mantelpiece. By acting on a "medium" scale scan such as this, we do not see the sensor drift present in hallway scans. Additionally this highlights how a large area can be scanned in high resolution, while we still see the detail in the individual busts.

\section{Conclusion}

In this paper, we presented a novel method for high resolution, spatially extended RGB-D mapping.  Our algorithm uses multiple mapping volumes balanced between the CPU and GPU to provide higher resolution than current state-of-the-art mapping systems.  We also presented a method for dynamically adding and removing mapping volumes to allow for spatially extended mapping.  We evaluated our method on multiple indoor and simulated datasets and showed results for a 48x increase in resolution over current state-of-the-art mapping. Furthermore, the dynamic placement of mapping volumes allows for the choice of mapping areas of interest, which can be object-centric or for full scene reconstruction.

\bibliographystyle{ieeetr}
\bibliography{./library}

\end{document}